\pdfoutput=1
\documentclass[11pt]{article}

\usepackage[]{EACL2023}

\usepackage{times}
\usepackage{latexsym}
\usepackage{enumitem}
\usepackage[T1]{fontenc}
\usepackage[utf8]{inputenc}

\usepackage{microtype}

\usepackage{inconsolata}

\usepackage{graphicx}

\usepackage[labelfont=bf]{caption}

\title{``Kurosawa'': A Script Writer's Assistant}

\author{Prerak Gandhi$^*$, 
  Vishal Pramanik$^*$,
  Pushpak Bhattacharyya \\
  Department of Computer Science and Engineering\\
  Indian Institute of Technology Bombay, Mumbai\\
\texttt{\{prerakgandhi,vishalpramanik,pb\}@cse.iitb.ac.in}}

\begin{document}
\maketitle
\def\thefootnote{*}\footnotetext{These authors contributed equally to this work}
\def\thefootnote{\arabic{footnote}}

\begin{abstract}
Storytelling is the lifeline of the entertainment industry-  movies, TV shows, and stand-up comedies, all need stories. A good and gripping script is the lifeline of storytelling and demands creativity and resource investment. Good scriptwriters are rare to find and often work under severe time pressure. Consequently, entertainment media are actively looking for automation. In this paper, we present an AI-based script-writing workbench called KUROSAWA which addresses the tasks of plot generation and script generation. Plot generation aims to generate a coherent and creative plot (600--800 words) given a prompt (15--40 words). Script generation, on the other hand, generates a scene (200--500 words) in a screenplay format from a brief description (15--40 words). Kurosawa needs data to train. We use a 4-act structure of storytelling to annotate the plot dataset manually. We create a dataset of 1000 manually annotated plots and their corresponding prompts/storylines and a gold-standard dataset of 1000 scenes with four main elements --- scene headings, action lines, dialogues, and character names --- tagged individually. We fine-tune GPT-3 with the above datasets to generate plots and scenes. These plots and scenes are first evaluated and then used by the scriptwriters of a large and famous media platform ErosNow\footnotemark\footnotetext{https://erosnow.com/}. We release the annotated datasets and the models trained on these datasets as a working benchmark for automatic movie plot and script generation. 
\end{abstract}
\section{Introduction}\label{sec1}
Movies are one of the most popular sources of entertainment for people worldwide and can be a strong medium for education and social awareness. The impact and influence of film industries can be gauged from the fact that Hollywood movies invest 100s of millions of dollars and often make box-office collections of billions of dollars. The first motion picture {\it The Great Train Robbery, 1903}--- black \& white with no sound--- was created at the beginning of the 20th century. Since then, the art has gone through several transformations, and now people can instantly access 4K HD movies of their liking on any smart device. 

Throughout the history of film, two of the contributors to a film's blockbuster success have been the quality of its plot and the manner of storytelling. The appeal of the movie decreases drastically if the viewers find the plot drably predictable. Writing a creative and exciting script is, therefore, a critical necessity and is extremely challenging. Add to this the constraints of time and budget, and the need for (at least partial) automation in script writing becomes obvious.

AI-based story generation has been used before. Based on the engagement-reflection cognitive explanation of writing, the computer model MEXICA \cite{perez2001mexica} generates frameworks for short tales. BRUTUS \cite{bringsjord1999artificial} creates short stories with predetermined themes like treachery. With the arrival of pre-trained transformer models, automatic story generation has got a shot in the arm. Transformer models like GPT-2 and GPT-3 are extensively used for text generation. These models have shown the capability of generating creative text, albeit sometimes with hallucinations \cite{zhao2020reducing}. Text generated by these models also sometimes lacks coherence and cohesiveness. On the other hand, template-based models can generate coherent text but lack creativity in generating new characters and events in the plot \cite{kale2020template}.

\begin{figure}
    \centering
    \includegraphics[width=1\columnwidth]{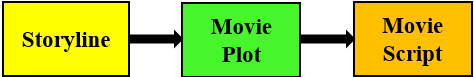}
    \caption{The thought process a scriptwriter follows in creating a movie script. An idea (\textbf{storyline}) leads to a \textbf{plot} which is then converted into a \textbf{movie script}.}
    \label{fig:movieprocess}
\end{figure}

The process of creating a movie generally starts with an idea which is then used to create a plot which is used as the base to build the movie script (Figure \ref{fig:movieprocess}).
 
Novel datasets are an important feature of this paper. We closely studied the plots and prompts of movies from Bollywood and Hollywood. Such plots and prompts were scraped from Wikipedia\footnotemark\footnotetext{\url{https://www.wikipedia.org/}} and IMDb\footnotemark\footnotetext{\url{https://www.imdb.com/}}, respectively. The plots are then annotated using the 4-act story structure- an extension of the well-known 3-act structure \cite{field1979screenplay}. The 4-act structure and the annotation methods are explained in detail in \textbf{appendix \ref{actdivision}} and \textbf{section \ref{sec5}}, respectively.

We introduce a dataset of 1000 Hollywood movie scenes and their short descriptions. The scripts are scraped from IMSDb\footnotemark\footnotetext{\url{https://www.imsdb.com/}}. The scenes are annotated with the four major components of a screenplay: {\it sluglines, action lines, character names} and {\it dialogues}, described in details in appendix \ref{screenplay}

We introduce a workbench which we call ``Kurosawa'', consisting of datasets and a pair of GPT-3 \cite{brown2020language} models fine-tuned with the said datasets. One GPT-3 model generates a movie plot given a short description of the storyline (15--40 words), while the other creates a scene based on a short description of the required scene. 

Importantly, we have provided the ``Kurosawa'' platform to one of the biggest media platforms engaged in the business of making movies and TV shows, producing music and soundtrack etc.- to help script and content writers from different film industries create new movie plots.

\vspace{3mm}
\textbf{Our contributions in this work are as follows:}
\begin{itemize}[noitemsep]
    \setlength\itemsep{0.1em}
    \item To the best of our knowledge, this is the first work on generating movie scenes from a scene description.
    \item We create and publicly release two datasets: (a) a parallel dataset of 1000 movie storylines and their corresponding plots, (b) a parallel dataset of 1000 movie scenes and their corresponding descriptions. In (a), we link available movie storylines from IMDb with available corresponding movie plots from Wikipedia. In (b), we link available movie scenes from IMSDb with corresponding descriptions from IMDb.
    \item We manually annotate movie plots according to a 4-act structure which is an extension of the well-known 3-act structure \cite{field1979screenplay}. Professional scriptwriters from the media and entertainment industry guided us very closely.
    \item We manually annotate movie scenes with four major components of a scene: \textit{sluglines, action lines, character names} and \textit{dialogues}, along with a short description of the scene.
    \item We introduce ``Kurosawa'': a workbench that consists of multiple datasets and models which can assist script and scene writers in the film industry.
\end{itemize}
\section{Motivation}\label{sec2}
Movies are a form of visual media and can have a huge influence on life and society. 
%The characters of a movie can influence the audience in both positive and negative ways. The youth and the children can learn a lot from the ideas and morals conveyed in a movie, and hence can have a long-lasting impression on society. 
Movie scripts are often 30,000 words long, comparable to a 100-page book. Though scripts can be diverse, they have fixed and oft-repeated structures, {\it e.g., scene heading, transition, character name, etc.}. This fixity and repetition can be dull and time-consuming and can be handed over to AI. However, a surprising fact is that AI-based models can be creative in generating novel characters and stories. These reasons have motivated the film industry to seriously consider harnessing AI for various aspects of movie making, script and scene writing being one of them.

Los Angeles Times, 19 December 2022, asks, "AI is here, and it's making movies. Is Hollywood ready?". The newspaper edition reports mainly movie editing efforts ongoing at various places using AI. Our task in the paper is allied but different in the sense that we aim to provide a "script-writers' assistant". 

%1. To reduce the time taken to generate a new idea for a movie script.
%2. Generate novel and creative plots.
%4. Generate interesting characters and their arcs.
%5. To provide writers with new ideas to progress their initial story by giving various options.
%6. It makes the process of making movie scripts cost-effective
\section{Related Work}\label{sec3}
%Text generation has always been an interesting area of research. There has been a considerable amount of work in story and plot generation, which is described in the following section. 

\subsection{Automatic Story Generation}
%The field of Natural Language Generation that is most closely related to movie plot and script generation is automatic story generation. There has been a plethora of work done in this field in the previous years. 
Neural models have been able to produce stories by conditioning on different contents like visuals \cite{huang2016visual} and succinct text descriptions \cite{jain2017story}. Work on plot controllable, plan-driven story generation abounds \cite{riedl2010narrative, fan2019strategies,perez2001mexica,rashkin2020plotmachines}. A related kind of work is automatic poetry generation based on keywords or descriptions \cite{yan2016poet,wang2016chinese}. 
%The task we present here is quite similar to the fields mentioned above, with the only difference being that, in our case, the number of words generated is higher. 

\subsection{Plot Generation}
Plot Machines \cite{rashkin2020plotmachines} generate multi-paragraph stories based on some outline phrases. \citet{fan2018hierarchical} introduce a hierarchical sequence-to-sequence fusion model to generate a premise and condition that in turn generate stories of up to 1000 words. This work- unlike ours- is non-neural and template-driven and is, therefore, much less creative and novel compared to what we generate. 

\subsection{Scene Generation}
Automatic scene or script generation has received comparatively less attention. Dialogue generation \cite{li2016deep,huang2018automatic,tang-etal-2019-target,wu-etal-2019-proactive} with a semblance of scene generation has been done. There has recently been some work focusing on guiding dialogues with the help of a narrative \cite{zhu-etal-2020-scriptwriter}. We generate scenes in which the main elements come from a small prompt as input.
\section{Dataset}\label{sec5}
For movie plot generation, we have taken the plots from Wikipedia. The prompts for this task have been taken from IMDb. In IMDb, this prompt can be of two types. The first is a short description (15--40 words) of the movie, while the second is a long storyline, which varies from 30--200 words and contains much more details about the different characters and events of the movie. We have also collected the genres of each film from IMDb. We then divide the plots using a 4-act structure. For scene generation, we take the scripts from IMSDb and annotate them with the key elements of a scene. 

\subsection{Plot Generation Dataset}
We have created a dataset of 1000 plots consisting of both Bollywood and Hollywood plots, extracted from Wikipedia using the \emph{wikipedia} module in python. The plots collected are around 700 words long on average.
%These word lengths are considered long texts for language models. We have annotated the plots to divide them into parts and help the model understand each part's nuances better.

%\subsubsection{Plot Cleaning}
%The plots collected are not good enough to use directly. Some plots have the actors' names in brackets next to the characters' names. This information is not helpful, and we manually removed it. Some of the plots are written in order of a timeline instead of sequential order, as depicted in the film and were removed. We also decided to be careful with sensitive content and discard plots with extreme sexual or abusive content. 

\subsubsection{Annotation Guidelines}
We annotate the plots by manually dividing them into 4 parts using the 4-act structure described in \textbf{appendix \ref{actdivision}}. We place a single tag at the end of each act: \textit{\textlangle{}one\textrangle{}} (Act 1),
\textit{\textlangle{}two-a\textrangle{}} (Act 2 Part A),
\textit{\textlangle{}two-b\textrangle{}} (Act 2 Part B) and 
\textit{\textlangle{}three\textrangle{}} (Act 3) as delimiters. An example for plot annotation is given in the \textbf{appendix (Figure \ref{fig:plotann})}.

\subsubsection{Movie Genres}
\begin{figure}
    \centering
    \includegraphics[width=1.0\columnwidth]{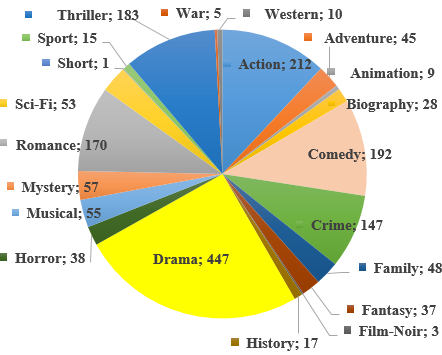}
    \caption{Genre distribution within the plot dataset}
    \label{fig:genregraph}
\end{figure}

To bring some controllability to the plots generated by the model, we have introduced the genres of the movies in the dataset along with the storyline. We concatenate the genres at the beginning of the storyline. Figure \ref{fig:genregraph} shows the distributions of genres in the dataset.  

% Need to improve the image clarity and aspect ratio.

\subsection{Scene Generation Dataset}
Movie scripts are very long. A 2-hour movie corresponds to around 30,000 words. Language models used for creative text generation, like GPT-2 and GPT-3, have token limits of 1024 and 2048, respectively, making it impossible to handle an entire script in one go. Hence, we divided the scripts into scenes and manually created their short descriptions. This allows training the scenes independently instead of relying on any previous scenes. 
%We gave the annotation task to external annotators owing to the large dataset. We have provided more information in section \ref{sec:annotator}.

Movie scripts comprise of multiple elements described in \textbf{appendix \ref{screenplay}}. The different elements increase the difficulty models face in learning to distinguish each element. To overcome this obstacle, we tag four major elements throughout the script --- \emph{sluglines, action lines, dialogues and character names}.  

\subsubsection{Annotation Guidelines}
\begin{figure}
    \centering
    \fbox{\includegraphics[width=\columnwidth]{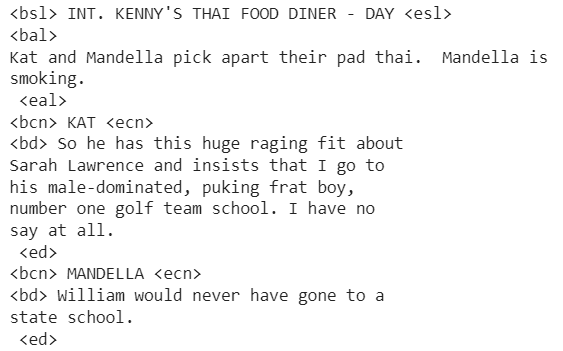}}
    \caption{The image depicts a portion of a movie scene with the four major elements annotated.}
    \label{fig:annot_script}
\end{figure}

We keep the four major elements present in every script --- \emph{sluglines, action lines, character name and dialogues}--- and remove any other type of information such as page number, transitions or scene dates. The tagging of the four major elements is done using beginning and ending tags that are wrapped around the elements, as shown below:
\begin{itemize}[noitemsep]
    \itemsep0em
    \item Sluglines: \textlangle{}bsl\textrangle{}...\textlangle{}esl\textrangle{}
    \item Action Lines: \textlangle{}bal\textrangle{}...\textlangle{}eal\textrangle{}
    \item Character Name: \textlangle{}bcn\textrangle{}...\textlangle{}ecn\textrangle{}
    \item Dialogue:\textlangle{}bd\textrangle{}...\textlangle{}ed\textrangle{}
\end{itemize}
An example of an annotated scene is seen in Fig \ref{fig:annot_script}.

\section{Experiments and Evaluation}
\label{Section 7}
We fine-tune GPT3 with our datasets (refer \textbf{appendix \ref{gpt3}}). 
\subsection{Plot Generation}
\label{sec7.1}
%The Hollywood plot dataset is used to create 2 models by fine-tuning GPT-2, (i) with annotation and (ii) without annotation.
We have created 5 models by fine-tuning GPT-3 with our movie plot dataset in the following manner, (i) \textbf{original} (without annotation) \textbf{(O)}: input- short storylines, output- plots without any annotations, (ii) \textbf{annotation and short input} \textbf{(AS)}: input- short storylines, output- plots annotated with 4-act structure, (iii) \textbf{annotation and long input} \textbf{(AL)}: input- long, more descriptive storylines, output- plots annotated with 4-act structure, (iv) \textbf{annotation and short input with genres included} \textbf{(ASG)}: input- short storylines and genre, output- plots annotated with 4-act structure, (v) \textbf{annotation and long input with genres included} \textbf{(ALG)}: input- long and more descriptive storylines along with the genre, output- plots annotated with 4-act structure.

For automatic evaluation we use \textbf{BLEU} \cite{papineni2002bleu}, \textbf{Perplexity} \cite{Jelinek1977PerplexityaMO}, \textbf{ROUGE} \cite{lin-2004-rouge}. We also use human evaluation in the form of a five-point Likert Scale \cite{alma991033023919703276}. The rating system has 1-> Strongly Disagree, 2-> Disagree, 3-> Neutral, 4-> Agree, 5-> Strongly Agree. Human-written stories are assumed to have a rating of 5 for each of the following 5 features: (1) \textbf{Fluency}: grammatical correctness; (2) \textbf{Coherence}: logical ordering of sentences and paragraphs; (3) \textbf{Relevance}: Whether the key points from the prompt have been highlighted in the output; (4) \textbf{Likability}: The measure of how much the story is enjoyable; (5) \textbf{Creativity}: If the output introduced any new events, character profiles, or relationships. 

For plot generation, we generate 50 plots from 50 test prompts. We divide the stories into five groups of 10 and assign three evaluators to each group. 

For scene generation, we generate ten scenes from 10 test prompts. We assign five evaluators to rate these ten stories.
\section{Results and Analysis}
We present our observations and evaluations. The nature of our task makes human evaluation take precedence over automatic evaluation (it is for automatic movie script generation, after all!). 
%Eros Now is an Indian subscription-based over-the-top, video-on-demand entertainment and media platform which offers media streaming and video-on-demand services. 
The qualitative analysis of our generated plots and scenes is based on feedback from 5 professional scriptwriters of our industry partner, the well-known media platform.

\subsection{Plot Generation}
\subsubsection{Automatic Evaluation}
Table \ref{tab:gpt3plotautoeval} shows auto-evaluation scores for the multiple GPT-3 plot generation models.

\begin{figure}[ht]
    \centering
\fbox{\includegraphics[width=1.0\columnwidth,height=0.8\columnwidth]{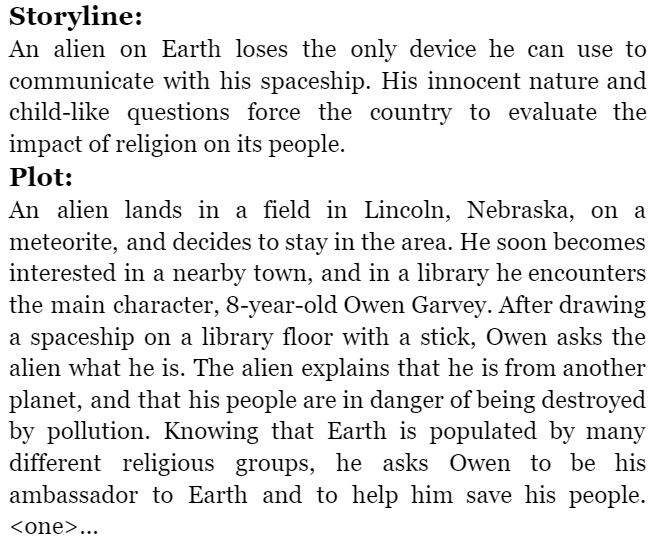}}
    \caption{The above paragraph is a partial example of a movie plot generated by the model fine-tuned with input as short storyline and output as plot annotated with the 4-act structure.}
    \label{fig:bestplotgen}
\end{figure}

\begin{table*}[htp]
% \small
% \resizebox{0.75\textwidth}{!}{%
    \centering
    \begin{tabular}{|l|l|l|l|l|l|}
        \hline
         \hfill \textbf{Models} & \textbf{O} & \textbf{AS} & \textbf{ASG} & \textbf{AL} & \textbf{ALG} \\
         \textbf{Metrics} & & & & & \\
         \hline
         \textbf{Perplexity} & 2.48 & \textbf{1.84} & 2.43 & 2.33 & 2.63\\ 
         \hline
         \textbf{BLEU-2 (\%)} & 12.95 & 12.01 & 12.51 & 13.08 & \textbf{14.52}\\ 
         \hline
         \textbf{BLUE-3 (\%)} & 4.70 & 4.21 & 4.55 & 4.84 & \textbf{5.59}\\ 
         \hline
         \textbf{BLUE-4 (\%)} & 2.14 & 1.92 & 2.13 & 2.27 & \textbf{2.59}\\ 
         \hline
         \textbf{ROUGE-L (\%)} & 22.67 & 21.72 & 23 & 24.02 & \textbf{24.88}\\
         \hline
         \textbf{Distinct 3-gram (\%)} & 97.55 & 97.61 & 97.39 & 97.28 & \textbf{98.09}\\ 
         \hline
         \textbf{Repetition 3-gram (\%)}& 1.99 & 2.02 & \textbf{1.72} & 1.89 & 1.74\\
         \hline
        \end{tabular}
        \caption{Scores from common evaluation metrics for 5 Hollywood plot generation models fine-tuned on GPT-3 as O, AS, ASG, AL, ALG (\ref{sec7.1}) }
        \label{tab:gpt3plotautoeval}
\end{table*}

\subsubsection{Human Rating}
\begin{figure*}[!h]
    \centering
   \fbox{\includegraphics[width=1.0\columnwidth,height=0.7\columnwidth]{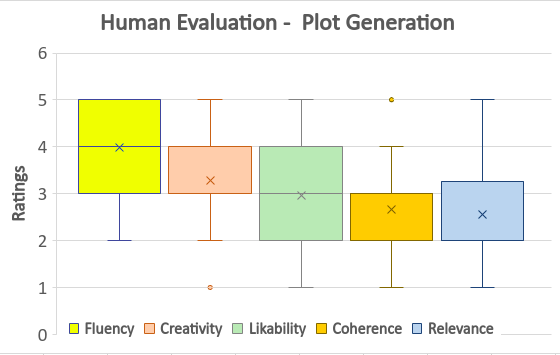}}
    \fbox{\includegraphics[width=1.0\columnwidth, height=0.7\columnwidth]{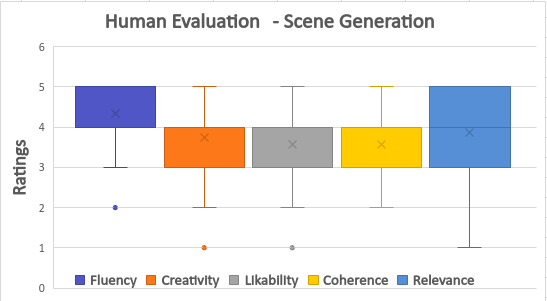}}
    \caption{Boxplot graphs for Human Evaluation of the plot and scene generation models.}
    \label{fig:humanevalplot}
\end{figure*}

We conducted human evaluation on Hollywood annotated short input model. The evaluation was done by five groups of 3 people, with each group having been assigned 10 unique plots. The ratings given for the 5 features are in Figure \ref{fig:humanevalplot}. The average scores for fluency, creativity, likability, coherence and relevance are \textbf{3.98}, \textbf{3.29}, \textbf{2.97}, \textbf{2.65} and \textbf{2.55}, respectively. Fluency of almost 4 is an indicator of the power of GPT-3 as a language model. Creativity and likeability are respectable at a value of around 3.0. The low BLEU scores support the average creativity score (Table \ref{tab:gpt3plotautoeval}). Figure \ref{fig:humanevalplot} indicates that coherence and relevance still have major room for improvement.

The MAUVE \citep{pillutla2021mauve} value measures the gap between neural text and human text. We have separately calculated the MAUVE scores for 20 plots and 50 plots. The weighted average of the MAUVE scores for the two experiments is \textbf{0.48} which is reasonably good. 

\subsubsection{Qualitative Observations}
Professional scriptwriters from our industry partner have given the following observations:
\newline

\textbf{Non-annotated Hollywood Plots}
\begin{itemize}[noitemsep]
    \itemsep0.0em
    \item The build-up is creative and interesting, but the ending becomes incoherent.
    \item Some characters which are introduced in the beginning are never mentioned again.
    \item The output is not portraying the key points or the theme mentioned in the input.
\end{itemize}

\textbf{Annotated Hollywood Plots}
\begin{itemize}[noitemsep]
    \itemsep0em
    \item The plots are much more coherent, and the endings are logical.
    \item There is still hallucination present (a common feature of all models).
    \item The longer inputs made the plots more attentive to the key points.
\end{itemize}

\textbf{Annotated Hollywood Plots with Genres included}
\begin{itemize}[noitemsep]
    \itemsep0em
    \item Along with the above points, now the plots generated are more tilted towards the genre or genres of the movie the writer wants to create.
    \item Addition of genre gives some control over the kind of plot generated by the model.
\end{itemize}

\textbf{Annotated Bollywood plots}
\begin{itemize}[noitemsep]
    \item The outputs show incoherence in the last two paragraphs and repetition of the same characters throughout the plot.
    \item The flow of the plot is not fast enough, i.e., the plot does not move ahead much.
    \item Many of the outputs have a 1990s theme around them, where the characters are separated and then find each other later. This is due to a skewed dataset with lesser modern plots.
\end{itemize}

\subsection{Scene Generation}
We fine-tuned GPT-3 for scene generation with our dataset. We generated ten scenes using the models mentioned in \ref{sec7.1}. Figure \ref{fig:scenegen} in the appendix. shows an example of a completely generated scene.

\subsubsection{Human Ratings}
We conducted a human evaluation on 10 scenes generated by the above model. 5 people evaluated the scenes using the Likert Scale. The ratings for the five features can be seen in Figure \ref{fig:humanevalplot}. The average scores for \textit{fluency, creativity, likability, coherence,} and \textit{relevance} are \textbf{4.48}, \textbf{3.9}, \textbf{3.48}, \textbf{3.46} and \textbf{3.86}, respectively. All of the values are above the neutral mark and imply that the generated scenes are close to human-written scenes.

\subsubsection{Qualitative Observations}
In this section, we analyze the quality of the scenes generated by the GPT-3 model. This analysis has been done by professional scriptwriters from the previously mentioned media company.
\begin{itemize}[noitemsep]
    \item The model produces a well-structured scene.
    \item It can create new characters and fabricate dialogues even when they are unimportant.
    \item The key points from the input can be found in the output.
    \item There are some lines that are repetitive.
    \item The output is not completely coherent.
\end{itemize}
\section{Conclusion and Future Work}
\label{sec9}
In this paper, we have reported a first-of-its-kind work on automatic plot and script generation from prompts. Automatic evaluation, human rating using the Likert scale, and qualitative observations by professional scriptwriters from our industry partner (a large and well-reputed media platform)- all vindicate the power of our rich dataset and GPT3 in script generation. We hope our work will help television show writers, game show writers, and so on.

There are several future directions: (i) the imbalance in the Bollywood plot dataset needs to be rectified; (ii) there is a lot of variation in Indian script because of multilingualism, which needs addressing; (iii) the most obvious weakness of GPT-3 is not being able to handle factual data and numbers, causing hallucination and preventing the automatic generation of documentaries and biographies. Detection and resolution of hallucination is anyway a growing need for language models. 
\section{Limitations}
\begin{itemize}[noitemsep]
    \itemsep0em
    \item In the plot generation dataset, the Wikipedia plots are sometimes not written by professional content writers from the film industry. Therefore these plots may fail to include the main events of the movie.
    \item In a few cases, the model fails to generate coherent events along with the abrupt introduction of characters in the plots and scenes.
    \item Although it has been noticed only a few times, the plot or scene generated contains repeated clauses or phrases.
    \item The model hallucinates and generates factually incorrect things, making it incapable of generating biographies or documentaries.
    \item The plot or scene may not abide by the theme of the input or genre mentioned along with the prompt.    
\end{itemize}
\label{sec10}
\bibliography{anthology,custom}
\bibliographystyle{acl_natbib}

\appendix
\section{Appendix}
\label{sec:appendix}
\subsection{Ethics Consideration}
\label{ethics}
We have taken all the scripts from IMDB and IMSDb databases. The
website has a disclaimer regarding using its scripts for research, which can be found at this link \url{https://imsdb.com/disclaimer.html}. We have used the scripts
fairly and without copyright violation. 

\subsection{Annotator Profiles}
\label{sec:annotator}
We required the help of external annotators in two cases: (i) Manually Annotating the Scripts and (ii) Creating scenes and their descriptions from the scripts. For the first task, we took the help of 10 annotators. Their ages ranged from 21-28, and all were Asian. They were given detailed guidelines with examples for annotating. There were also periodic sessions to confirm their understanding and solve their doubts and mistakes. For the second task, we took the help of two annotators. Both of them are Asian females aged between 21-23. Both of them were given detailed guidelines for the scene-writing task. A few data points were picked randomly and checked to find out and correct conceptual mistakes. The annotators had bachelors and masters degree in STEM and Arts.

\begin{figure*}[h]
    \centering
    \fbox{\includegraphics[width=0.8\textwidth, height=0.6\textwidth]{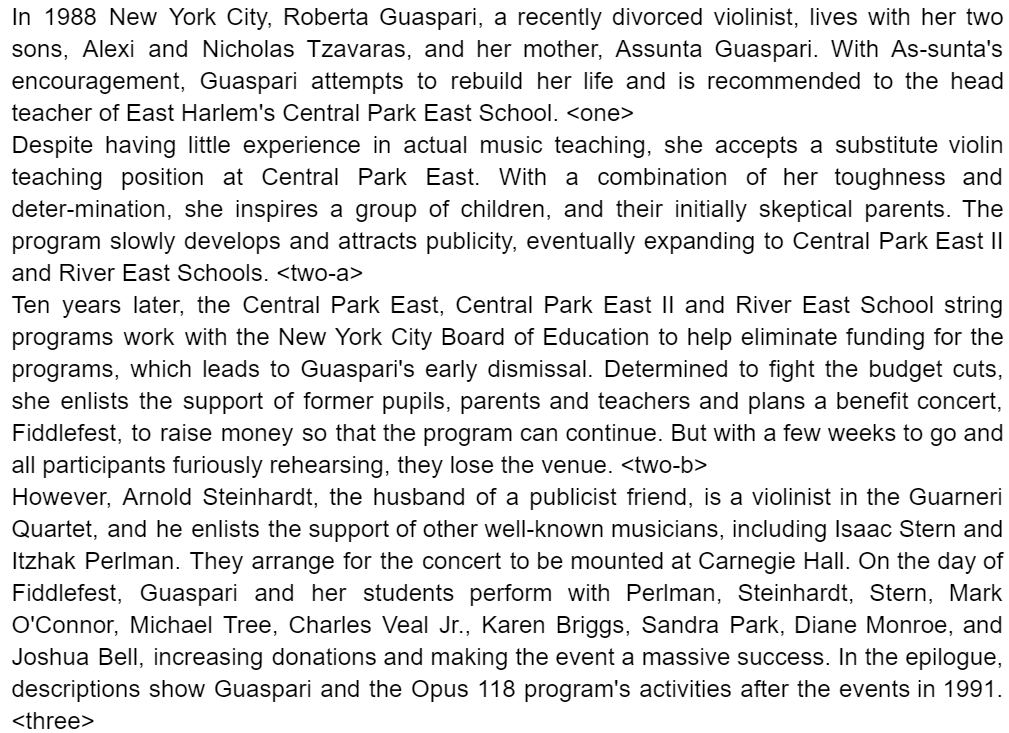}}
    \caption{Example of manual annotation of the plot of the movie \emph{Music of the Heart} using the 4-act structure}
    \label{fig:plotann}
\end{figure*}

% \subsection{Scene Generation Example}
% \label{sec:sceneexamples}
\begin{figure*}[h]
    \centering    \fbox{\includegraphics[width=0.95\textwidth,height=1.0\textwidth]{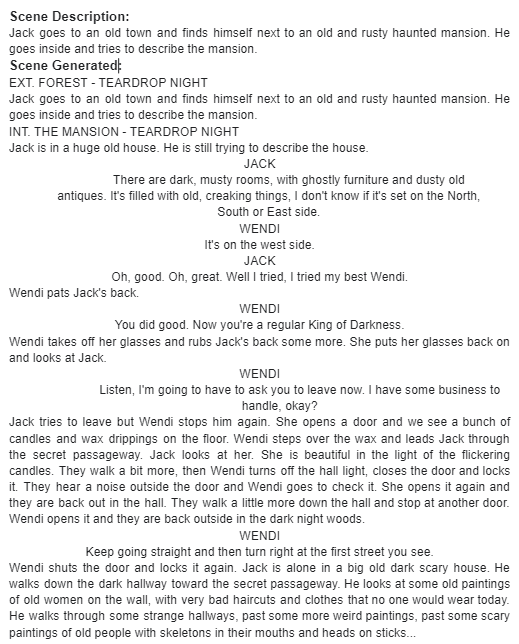}}
    \caption{An example of a complete scene generated given a short input.}
    \label{fig:scenegen}
\end{figure*}

% \subsection{Plot Generation Examples}
% \label{sec:plotexamples}
% \begin{figure}[h]
%     \centering
%     \fbox{\includegraphics[width=1.0\textwidth]{EMNLP 2022/Sections/PKPlot.PNG}}
%     \caption{An example of a complete plot generated given a storyline.}
%     \label{fig:plotcompletegen}
% \end{figure}

% \subsection{Steps for Finetuning GPT-3}
% \label{sec:gpt3appendix}

% GPT-3 was released by OpenAI\footnotemark\footnotetext{\url{https://openai.com/}} for public use in 2021. It consists of various pretrained models, with the largest model having 175 billion parameters. The models are not downloadable but can be accessed through API following account creation. The method to create datasets and to fine-tune the GPT-3 models are available at \url{https://beta.openai.com/docs/guides/fine-tuning}. We will, nevertheless, explain the steps required to fine-tune a GPT-3 model.
\subsection{Evaluation Metrics}
\label{metrics}
The evaluation metrics are described below:
\begin{itemize}
    \item \textbf{Perplexity (PPL)}: Perplexity is one of the most common metrics for evaluating language models. They are computed as exponential of entropy. The smaller the value of the PPL, the greater the fluency of the generated text.
    \item \textbf{BLEU}: \textbf{B}i\textbf{L}ingual \textbf{E}valuation \textbf{U}nderstudy is a common metric in many NLP tasks, especially in the field of Machine Translation. It measures the overlap between the generated output and gold standard data. Although this metric does not consider the model's creativity, we can deduce the difference between the candidate text and the reference text using BLEU. The higher the BLEU measure, the better it is.
    \item \textbf{ROUGE}: \textbf{R}ecall-\textbf{O}riented \textbf{U}nderstudy for \textbf{G}isting \textbf{E}valuation is typically used for evaluating automatic summarization. In our case, it measures the longest overlapping sequence between the generated and original plots. The higher the ROUGE measure, the better it is.
    \item \textbf{N-grams}: We measure the redundancy and diversity of the movie plots by computing the repetition and distinction n-gram scores.
\end{itemize}

%\subsection{Background and Terminologies}
%\label{sec4}
%In this section, we will explore the structure of a screenplay and some story templates in detail.

\subsection{Screenplay Structure}
\label{screenplay}
A movie script or a screenplay has a different format than a story. A script is a group of scenes. Each of these scenes consists of a few major components, which are discussed below: 

\textbf{Scene Headings/Sluglines}- This component describes the when and where of the scene. It can be thought of as the first shot that a camera takes of a new scene. For example, INT. - RESTAURANT - NIGHT indicates that the scene starts inside a restaurant at night. Sluglines are normally written in capital letters and are left-aligned.

\textbf{Character Names}- they are mentioned every time a character is going to utter a dialogue. The name of each character is mentioned in uppercase and is centre aligned.

\textbf{Dialogues}- dialogues are the lines that the characters say. They appear right after the character name in a script and are centrally aligned.

\textbf{Action Lines}- action lines describe almost everything about a scene. They can be described as the narration of each script. Action lines can be present after either dialogues or sluglines and are left-aligned.

\textbf{Transitions}- a transition marks the change from one scene to the next. They also depict how a scene is ended. For example, DISSOLVE, FADE, and CUT are different keywords used to indicate a transition. They are usually in upper case and are right-aligned. 

\begin{figure*}[h]
    \centering
    \fbox{\includegraphics[width=1\textwidth, height=0.6\textwidth]{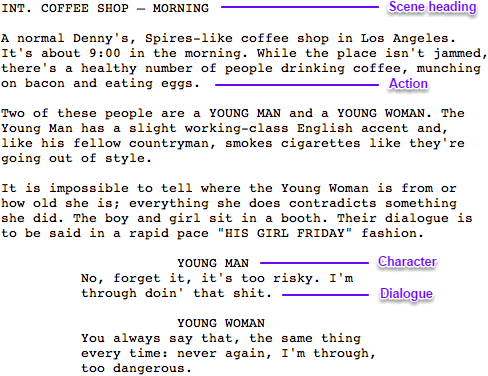}}
    \caption{The elements of a screenplay}
    \label{fig:screenplay}
\end{figure*}

Figure \ref{fig:screenplay} shows an example of the screenplay elements.

\subsection{Story Templates}
\label{actdivision}

%Storytelling has been prevalent as a means of human communication for thousands of years
Over time various templates have been developed that help to create stories. One of the most famous templates is the 3-act structure \cite{field1979screenplay}. This structure divides a story into a \textit{setup, confrontation, and resolution}. In this work, we have used the 4-act structure which we now describe in detail.

\textbf{Act 1}- This is the opening/introduction act. It describes the protagonist's character and briefly introduces the movie's theme. The act ends with the start of a new journey for the protagonist.

\textbf{Act 2A}- Due to the vast span of Act 2, it can be divided into two acts. This act usually contains the start of a love story. It also entertains the audience as the protagonist tries to adapt to their new journey. The act ends as the movie's midpoint, one of the film's critical moments, with either a very positive or negative scene.
    
\textbf{Act 2B}- This act usually contains the protagonist's downfall. The villain or antagonist starts to gain an advantage, and the protagonist loses something or someone significant. The act ends with the protagonist realizing their new mission after reaching rock bottom.

\textbf{Act 3}--- The protagonist has realized the change required in them and sets out to defeat the antagonist in a thrilling finale. The movie then ends by displaying a welcome change in the protagonist that was lacking in the beginning.

\subsection{Fine-Tuning GPT-3}
\label{gpt3}
GPT-3 was deemed publicly available last year by OpenAI \cite{brown2020language}. Its best model has 175B parameters, which is much more than GPT-2's 2.9B parameters. We have fine-tuned multiple plot generation models with GPT-3 along with a scene generation model. The multiple combinations of plot generation models are short or long prompts and with or without genres. The GPT-3 model and hyperparameters remain the same for all the above combinations. We have fine-tuned the GPT-3 Curie model for four epochs. For generating text, GPT-3 offers various hyperparameters to tune and get closer to our desired results. For testing, we set other hyperparameters as follows: the temperature as 0.7, top-p as 1, frequency penalty as 0.1, presence penalty as 0.1, and max tokens as 900.

\end{document}